\documentclass[runningheads]{llncs}
\usepackage[T1]{fontenc}
\usepackage{graphicx}
\usepackage{url}
\usepackage{times}
\usepackage{soul}
\usepackage[hidelinks]{hyperref}
\usepackage[small]{caption}
\usepackage{graphicx}
\usepackage{amsmath}

\usepackage{amsthm}
\usepackage{booktabs}
\usepackage{amsmath,tabu}
\usepackage{amssymb,amsfonts,textcomp,dsfont}
\usepackage{verbatim}
\usepackage{multicol}
\usepackage{tabularx}

\usepackage{subfig}
\newcommand{\weak}{\text{:\texttildelow} \ }
\usepackage{textcomp}
\newcommand{\clingo}{\textsf{\footnotesize Clingo}}
\newcommand{\asal}{\textsf{\footnotesize ASAL}}
\newcommand{\rpni}{\textsf{\footnotesize RPNI}}
\newcommand{\edsm}{\textsf{\footnotesize EDSM}}
\newcommand{\disc}{\textsf{\footnotesize DISC}}
\newcommand{\obs}{\textsf{\scriptsize obs}}
\newcommand{\av}{\textsf{\scriptsize av}}

\theoremstyle{definition}
\newtheorem{mydef}{Definition}

\usepackage{algorithm}
\usepackage[noend]{algpseudocode}
\algnotext{EndFor}
\algnotext{EndIf}

\begin{document}
\title{Learning Automata-Based Complex Event Patterns in Answer Set Programming}
%
%\titlerunning{Abbreviated paper title}
% If the paper title is too long for the running head, you can set
% an abbreviated paper title here
%
%\author{Nikos Katzouris\inst{1} \and Georgios Paliouras\inst{1}}
\author{Nikos Katzouris \and Georgios Paliouras}
\authorrunning{Katzouris \& Paliouras}
% First names are abbreviated in the running head.
% If there are more than two authors, 'et al.' is used.
%
\institute{Institute of Informatics \& Telecommunications, National Center for Scientific Research (NCSR) ``Demokritos'', Athens, Greece\\
	\email{\{nkatz,gpaliourg\}@iit.demokritos.gr},\\ 
	%WWW home page:
	%\texttt{http://users/\homedir iekeland/web/welcome.html}
	%\and
	%Universit\'{e} de Paris-Sud,
	%Laboratoire d'Analyse Num\'{e}rique, B\^{a}timent 425,\\
	%F-91405 Orsay Cedex, France
}
\maketitle              % typeset the header of the contribution
\begin{abstract}
Complex Event Recognition and Forecasting (CER/F) techniques attempt to detect, or even forecast ahead of time, event occurrences in streaming input using predefined event patterns. Such patterns are not always known in advance, or they frequently change over time, making machine learning techniques, capable of extracting such patterns from data, highly desirable in CER/F. Since many CER/F systems use symbolic automata to represent such patterns, we propose a family of such automata where the transition-enabling conditions are defined by Answer Set Programming (ASP) rules, and which, thanks to the strong connections of ASP to symbolic learning, are directly learnable from data. We present such a learning approach in ASP and an incremental version thereof that trades optimality for efficiency and is capable to scale to large datasets. We evaluate our approach on two CER datasets and compare it to state-of-the-art automata learning techniques, demonstrating empirically a superior performance, both in terms of predictive accuracy and scalability.

\end{abstract}
\section{Introduction}
Complex Event Recognition and forecasting (CER/F) systems \cite{cugola2012processing,DBLP:journals/vldb/GiatrakosAADG20,alevizos2022complex} detect, or even forecast ahead of time, occurrences of \emph{complex events} (CEs) in multivariate streaming input, defined as temporal combinations of \emph{simple events}, e.g. sensor data. CE patterns are typically defined by domain experts in some \emph{event specification language}. However, such patterns are not always known in advance, or they frequently need to be revised, as the characteristics of the input data change over time, making machine learning techniques capable of learning such patterns from data highly desirable in CER/F.

Despite the great diversity of existing event specification languages, a minimal set of basic constructs/operators that should be present in every such language have been identified \cite{zhang2014complexity,alevizos2017probabilistic,DBLP:journals/vldb/GiatrakosAADG20,grez2019formal}, in the form of an \emph{abstract event algebra}. %Given such a set of operators, event patterns may be specified compositionally, by combining such operators together. 
The most important of these operators are the \emph{sequence operator} and the closely related \emph{iteration operator (Kleene Closure)}, implying respectively that some particular events should succeed one another temporally, or that an event should occur iteratively in a sequence, and the \emph{selection} operator, which filters (selects) events that satisfy a set of predefined predicates. Taken together, these three operators already point to a computational model for CER/F based on symbolic automata \cite{d2017power}, and indeed, in most existing CER/F systems CE patterns are either defined directly as symbolic automata, or are compiled into such at runtime \cite{wu2006high,agrawal2008efficient,zhang2014complexity,diao2007sase+,demers2006towards,demers2007cayuga,schultz2009distributed,pietzuch2003framework,cugola2010tesla,alevizos2022complex}. In symbolic automata the transition-enabling conditions are predicates, rather than mere symbols, as in classical automata. In CER/F, the structure of such an automaton pattern corresponds to the conditions specified by the sequence/iteration operators in the CE pattern and their transition guards correspond to the pattern's selection predicates.
%Add refs for FlinkCEP and Sidhi.

Learning such symbolic automata-based CE patterns is a challenging task that requires to combine automata structure identification techniques with reasoning about the satisfiability of the selection predicates. To address this issue we propose a family of symbolic automata, which we call \emph{answer set automata (ASA)}, where the transition guards are defined by means of rules in Answer Set Programming (ASP), providing definitions for the selection predicates. Importantly, thanks to the strong connections of ASP with symbolic learning, ASA-based CE patterns are directly learnable from data. We present an approach that utilizes the power of ASP in declarative learning to automatically construct such patterns from data and we lift this approach to an incremental version that trades optimality for efficiency and allows our learning framework to scale to large datasets. We evaluate our approach on two CER datasets and compare it to state-of-the-art automata learning techniques, demonstrating empirically a superior performance, both in terms of predictive accuracy and scalability.

\section{Related Work}
   
Although the field of automata learning \cite{de2010grammatical,wieczorek2017grammatical} has a long history in the literature \cite{gold1967language,angluin1987learning,oncina1992identifying,lang1998results,ulyantsev2015bfs,angluin2015learning,giantamidis2021learning,smetsers2018model,furelos2021induction}, existing techniques that induce automata from positive \& negative traces have several shortcomings, which limit their applicability in the CER/F domain. Most such algorithms either attempt to learn a model that perfectly discriminates between the positive/negative traces \cite{gold1967language,angluin1987learning,giantamidis2021learning,angluin2015learning,furelos2021induction}, or they use greedy techniques for \emph{state merging} \cite{oncina1992identifying,lang1998results}, %a form of agglomerative clustering that seeks to generalize from 
a technique that generalizes from a large, tree-like automaton (the \emph{Prefix Tree Acceptor -- PTA}), generated from the entire training set. % prefixes of the traces in the training set, by merging together states that exhibit similar behavior w.r.t. acceptance/rejection of the training data. 
These approaches tend to learn large, overfitted models that generalize poorly. % and are hard to inspect and interpret. 
More recent techniques \cite{shvo2021interpretable} replace the PTA generalization heuristics with exact, constraint-based automata identification methods, achieving higher generalization capacity. %Although DISC achieves better results as compared to classical state-merging, 
However, the issue remains that such techniques still need to encode the entire training set into a PTA, raising memory/scalability issues in large datasets. %For instance, training data in INFORE's biological and maritime use-cases consist of hundreds of thousands of traces and are well out of reach for existing automata induction techniques.

All aforementioned algorithms learn classical automata. Although some of these algorithms could, in principle, be applied to the symbolic automata learning setting, e.g. via propositionalization, that would entail an explosion in alphabet size and the combinatorial complexity of the learning task. On the other hand, although some algorithms for symbolic automata induction do exist \cite{drews2017learning,maler2017generic,argyros2018learnability,fisman2021inferring}, they are mostly based on ``upgrading'' existing classical automata identification techniques to richer alphabets, and they thus suffer from the limitations outlined above, i.e. poor generalization, intolerance to noise and limited scalability.

Learning automata and grammars has been an application domain for ILP since its early days, targeting mostly classical automata expressed as definite clause grammars. More recent ILP frameworks, such as meta-interpretative learning (MIL) and learning from answer sets (ILASP), have also been applied to the task \cite{muggleton2014meta,furelos2021induction}. However, both these approaches learn models that perfectly discriminate between positive and negative examples, therefore, they cannot deal with noise in the data. Moreover, the MIL approach of \cite{muggleton2014meta} learns classical automata, and although the ILASP-based approach of \cite{furelos2021induction} does learn a form of symbolic automata, the transition guards therein are restricted to propositional clauses generated from combinations of the alphabet symbols, which falls short of the CER/F requirement for arbitrary selection predicates.

In contrast to the above-mentioned approaches, our symbolic automata learning framework utilizes the full expressive power of ASP in the definitions of transition guards and ASP's declarative learning capabilities to learn highly compressive models that generalize adequately. Moreover, its incremental learning version is able to scale to arbitrarily large datasets.

%In contrast to the above our ASP-based approach is able to learn expressive event patterns in the form of symbolic automata, generalizing adequately and scaling to large datasets, thanks to its incremental learning strategy.

\section{ASP Background}
\label{sec:background}
We assume some familiarity with ASP and refer to \cite{lifschitz2019answer} for an in-depth account. In this section we review some basic ASP constructs that will be useful in what follows. Throughout, we use the \clingo\footnote{\url{https://potassco.org/}} syntax for representing ASP expressions. A choice rule is an expression of the form $\{\alpha\} \leftarrow \delta_1,\ldots,\delta_n$, %which is syntactic sugar for $a \leftarrow \delta_1,\ldots,\delta_n, \mathtt{not \ not} \ a$, and 
with the intuitive meaning that whenever the body $\delta_1,\ldots,\delta_n$ is satisfied by an answer set $I$ of a program that includes the choice rule, instances of the head $\alpha$ are arbitrarily included in $I$ (satisfied) as well. %A constraint is a rule $\leftarrow \delta_1,\ldots,\delta_n$  with an empty head. %, which is syntactic sugar for $\mathtt{false} \leftarrow \delta_1,\ldots,\delta_n$. 
%and means that the body cannot be satisfied. 
A weak constraint is an expression of the form $\weak \delta_1,\ldots,\delta_n. [w@p,t_1,\ldots,t_k]$, where $\delta_i$'s are literals, called the body of the constraint, $w$ and $p$ are integers, called respectively the \emph{weight} and the \emph{priority level} of the constraint and $t_1,\ldots,t_k$ are ASP terms. A grounding/instance of a weak constraint $c$ is an expression that results from $c$ by replacing all variables in $\delta_1,\ldots,\delta_n,t_1,\ldots,t_k$ by constants. Such an instance is satisfied by an answer set $I_{\Pi}$ of a program $\Pi$ that includes $c$ if $I_{\Pi}$ satisfies $c$'s ground body, which incurs a penalty of $w$ on $I_{\Pi}$. $I_{\Pi}$'s total cost is the sum of penalties resulting from each instance of $c$ that is satisfied by $I_{\Pi}$. Inclusion of weak constraints in an ASP program triggers an optimization process that yields answer sets of minimum cost. Priority levels in  weak constraints model the constraints' relative importance, since the aforementioned optimization process attempts to first minimize the total cost due to weak constraints of higher priority levels.

\section{The Problem Setting and a Running Example}

We next set the scene for our proposed automata-based CE pattern learning framework. CER/F applications usually deal with multivariate input, i.e., input that arrives in multiple streams, each representing a ``signal'' obtained by the evolution of a relevant domain attribute in time. We illustrate the case using a running example from the domain of precision medicine, as formulated in the context of the INFORE EU-funded project\footnote{\url{https://www.infore-project.eu/}}.

%An example that typically occurs in CER/F domains concerns multivariate time series data, which may be discretized and converted into symbolic sequences. Simple events in such cases may be defined as observations drawn directly from the data, e.g. some domain attribute ranging in a particular value interval, in addition to value aggregates and/or boolean combinations thereof. In turn, complex events are defined in FSA-based patterns that combine such simple events over time. 

\begin{figure}[t]
	\centering
	\includegraphics[width=0.6\textwidth]{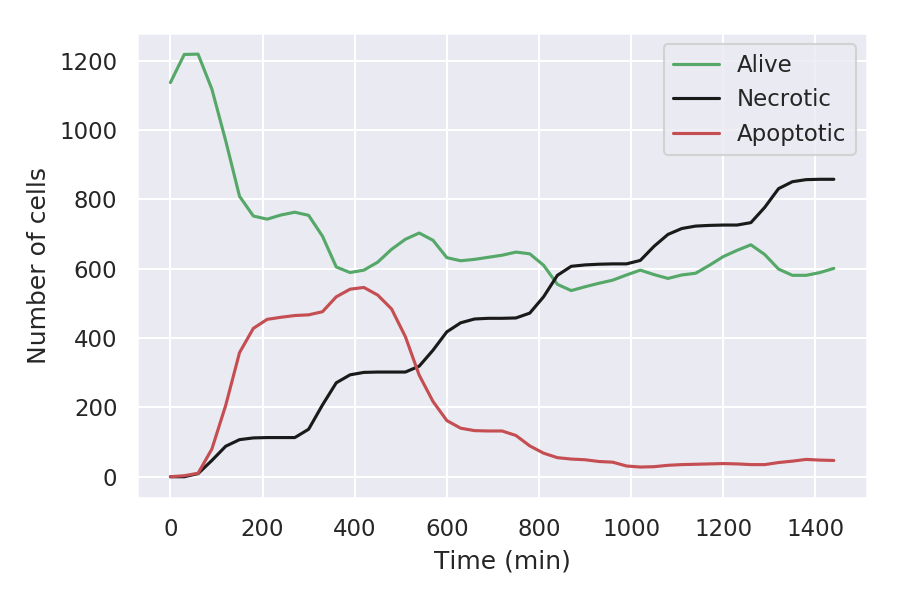}
	%\vspace*{-0.6cm}
	\caption{A multi-cellular simulation of tumor evolution from the INFORE project.}
	\label{fig:simulation}
\end{figure}

\begin{example}[\textbf{Running Example}]
	\label{ex:running}
	INFORE's precision medicine use-case utilizes CER/F techniques to assist the knowledge discovery process in personalized cancer treatment development. This process involves running a vast amount of complex, multi-cellular simulations to study the effects of various drug synergies on tumor growth. Such simulations are extremely demanding computationally and their majority ends-up in a negative result, signifying that a particular drug combination is not effective. Using CER/F to detect/forecast non-promising simulations at an early stage may thus speed up research by allowing to terminate non-interesting simulations early-on and allocate resources towards the exploration of more promising drug combinations. This calls for learning patterns of interesting/non-interesting outcomes from labeled, historical simulation data. %, in order to use such patterns with a CER/F system on new simulations. 
	Notably, the interpretability of such patterns is crucial in this domain.%, where the medical personnel needs to be able to trace and comprehend the system's predictions. 
	Therefore, mainstream black-box time-series classification methods, including deep learning techniques, are not an option. Figure \ref{fig:simulation} presents such a simulation generated by PhysiBoss \cite{letort2019physiboss}, a bio-informatics simulation environment that allows to explore the results of several environmental and genetic alterations to populations of cells. Figure \ref{fig:simulation} presents tumor growth evolution over time, in terms of population sizes of three types of cell: the tumor's \emph{alive} cells, its \emph{apoptotic} cells, i.e. cells that are ``programmed'' to die, due to apoptosis, and the tumor's \emph{necrotic} cells, i.e. cells that die due to the effects of an injected Tumor Necrosis Factor (TNF), i.e. a drug combination. %\hfill $\blacksquare$
\end{example}

\noindent Multivariate time-series, such as the simulation data from Figure \ref{fig:simulation}, may be converted into symbolic multivariate sequences (MVSs), e.g. by using the Symbolic Aggregate Approximation (SAX) algorithm  \cite{lin2003symbolic,lin2007experiencing}. SAX converts time-series into symbolic sequences, by mapping numerical values to symbols, drawn from a fixed-length alphabet, such that each symbol in the converted sequence corresponds to a bin (value range) in the original time-series. 

In the event pattern learning setting that we put forward we assume that the training data consist of labeled, symbolic MVSs, each representing a training example. For instance, the symbolic MVS obtained by discretizing the simulation data in Figure \ref{fig:simulation}, along with a label (e.g. \emph{interesting/non-interesting} simulation) represents such a training example. In what follows, by MVS we always mean a symbolic MVS.

Given an MVS $S$ of maximum length $n$ (i.e. the length of the largest sequence in $S$) we use a logical representation of $S$ as a sequence of interpretations $I_1,\ldots,I_n$, where each $I_t$ consists of ground facts that describe $S$'s $t$-th coordinate. In particular, we assume that each sequence in $S$ corresponds to a domain attribute and consists of symbols from a fixed alphabet $\Sigma$ that represent the values of this attribute over time. Then, the interpretation $I_t$ that corresponds to $S$'s $t$-th coordinate consists of \emph{observation facts} of the form $\mathsf{obs}(SeqId,\mathsf{av}(A,V),T)$, meaning that the attribute $A$ of the MVS with unique id $SeqId$ has value $V$ at time $T$.

\begin{table*}[t]
	%\footnotesize
	\begin{minipage}{\textwidth}
		\begin{tabular}{ p{0.5\textwidth} p{0.5\textwidth}} 
			\toprule
			\textbf{\emph{(a) A toy training set consisting of two MVSs.}} & ~\\
			\textbf{Positive example ($id_1$):} & \textbf{Negative example ($id_2$):}\\
			\emph{Alive cells:} \  \ \ \  \  \ \ \  \texttt{eeeedcbbbb}  &  \emph{Alive cells:} \  \ \ \  \  \ \ \   \texttt{eecdbbbbbb}\\
			\emph{Necrotic cells:} \ \   \texttt{aabbbcccde} &  \emph{Necrotic cells:} \ \    \texttt{aabbbbcccc}\\
			\emph{Apoptotic cells:} \ \texttt{bbbcdghhhh} & \emph{Apoptotic cells:} \  \texttt{bbbcfghhhh}\\ \\
			\textbf{\emph{(b) The logical representation of MVS $id_1$:}} & ~ \\
			$I^{id_1}_{1} = \{\obs(id_1,\av(alive,\texttt{e}),1). \ \obs(id_1,\av(necrotic,\texttt{a}),1). \ \obs(id_1,\av(apoptotic,\texttt{b}),1).\}$ & ~ \\
			$I^{id_1}_{2} = \{\obs(id_1,\av(alive,\texttt{e}),2). \ \obs(id_1,\av(necrotic,\texttt{a}),2). \ \obs(id_1,\av(apoptotic,\texttt{b}),2).\}$ & ~ \\
			$I^{id_1}_{3} = \{\obs(id_1,\av(alive,\texttt{e}),3). \ \obs(id_1,\av(necrotic,\texttt{b}),3). \ \obs(id_1,\av(apoptotic,\texttt{b}),3).\}$ & ~ \\
			\ldots & ~ \\
			$I^{id_1}_{10} = \{\obs(id_1,\av(alive,\texttt{b}),10). \ \obs(id_1,\av(necrotic,\texttt{e}),10). \ \obs(id_1,\av(apoptotic,\texttt{h}),10).\}$ & ~ \\
			%\hline
			\midrule
		\end{tabular}\\
	\end{minipage}
	%\vspace*{-0.3cm}
	\caption{Multivariate, discrete sequences and their logical representation.}
	\label{table:input-data}
\end{table*}
\normalsize

\begin{example}[\textbf{Labeled MVS}]
	\label{ex:input-data}
	Table \ref{table:input-data} (a) presents a minimal, toy training set with a single positive and a single negative example. Each example is an MVS consisting of three length-ten sequences regarding the evolution of cell populations for the different types of cell in our running example (Example \ref{ex:running}). These sequences are short excerpts of longer simulation sequences, which have been discretized using SAX. Each symbol in a sequence corresponds to a bin of real values.%, such that the lexicographic ordering between the symbols corresponds to the standard ordering between the bins (i.e. ``larger'' letters correspond to bins of larger values). 
	Table \ref{table:input-data} (b) presents the logical representation of the positive example  (MVS $id_1$) as a sequence of interpretations $I^{id}_1,\ldots,I^{id}_{10}$. %\hfill $\blacksquare$%, as per Definition \ref{def:training_exmpl}. The MVS interpretation $I_{id_1}$ is defined as the union of its coordinates, as per Definition \ref{def:coordinates}. There are ten coordinates, one for each position index in the original sequences in the MVS, with the $I_{id_1}^{t}$-th coordinate containing sequence information ($seq/3$ facts) regarding the $j$-th position in the sequences. The alphabet of the MVSs is $\Sigma = \{\mathtt{a,b,c,d,e,f,g,h}\}$ and the set $P$ of predicates symbols corresponding to the sequences' attributes ``names'' of Definition \ref{def:training_exmpl} is $P = \{alive, necrotic, apoptotic\}$.% and the extended alphabet $\Sigma^{+}$ of Definition \ref{def:training_exmpl} is $\Sigma^{+} = \{p(x) | \ p \in P, x\in \Sigma \}$.  
\end{example}

\section{Answer Set Automata}
\label{sec:asa} 

\noindent We next define a family of symbolic automata that may be used to express CE patterns over MVSs. %Symbolic automata are an extention of classical automata, where the transitions are annotated by predicates, rather than mere symbols. These predicates act as \emph{transition guards}, i.e. transition-enabling conditions. 
The transition guard predicates, which we shall call \emph{transition features} and correspond to selection predicates in a CER/F context, will be defined by ASP rules, we therefore call the resulting automata \emph{answer set automata} (ASA). Intuitively, a transition in an ASA is enabled when the body of the corresponding transition feature rule is satisfied by the input MVS. Note that ASA will be non-deterministic in principle. This is because to enforce determinism in the case of symbolic automata it must be ensured that the conditions that guard all outgoing transitions from some state $q$ to a set of different states must be mutually exclusive. This is infeasible to guarantee in the ASA framework, since the transition features may encode arbitrary, domain-specific conditions that are deemed informative to synthesize automata with. The non-determinism of the ASA framework is in accordance with most event specification languages in the CER domain, where complex event patterns are defined as, or are eventually compiled into non-deterministic automata.% [REFS: SASE, SASE+, FlinkCEP, ESPER, T-REX,  Cayuga [45, 46 from Alevizos]]. We next formally define ASA.

\begin{mydef}[\textbf{Answer Set Automaton -- ASA}]
	\label{def:asa}
	An answer set automaton is a tuple $\mathcal{A} = \langle \Sigma, B, R, Q, q_0, F, \delta  \rangle$, where: $\Sigma$ is the alphabet, represented by a set of ASP facts; $B$ is some background knowledge represented by an ASP program; $R$ is a set of ASP rules, called the \emph{transition features}; $Q$ is a finite set of \emph{states}; $q_0\in Q$ is a designated \emph{start state} and $F\subseteq Q$ is a designated set of \emph{accepting states}; $\delta$ is a non-deterministic state transition function defined by means of a \emph{feature mapping} $\delta_{R}: Q\times R \rightarrow Q$, where $\delta_{R}(q,r) = q'$ has the intuitive meaning that the transition from state $q$ to $q'$ is guarded by the transition feature $r$. Given a feature mapping $\delta_{R}$, the transition function $\delta: Q\times 2^{\Sigma} \rightarrow 2^Q \cup \{\bot\}$ is defined as: 
	
	\begin{equation}
	\label{eq:delta}
	\delta(q,I)= \left\{
	\begin{array}{ll}
	A_{q} = \{\delta_{R}(q,r)\in Q \ | \ I\cup B \vDash body(r)\} \text{, if } A_{q}\neq \emptyset, \\
	\square\in \{\{q\},\{\bot\}\}, \text{ else.}\\
	\end{array} 
	\right. 
	\end{equation}
\end{mydef}  

%\begin{remark}
	\noindent The ``alphabet'' $\Sigma$ in Definition \ref{def:asa} is defined as the set of ground facts that encode the simple events in the input, i.e., facts of the form $\mathsf{obs}(alive,\mathtt{a})$, as per our running example. The transition function $\delta$ operates on interpretations, i.e. subsets of $\Sigma$, as indicated by the powerset $2^{\Sigma}$ in $\delta$'s signature. Then, given a state $q$ and an interpretation $I$, the ``if'' branch of $\delta$ in Eq. (\ref{eq:delta}) maps $q$ to its set of ``next states'' $A_{q}$. Each $q' \in A_{q}$ is obtained by the feature mapping $\delta_{R}$, therefore, it is of the form $q' = \delta_{R}(q,r)$ and has the property that $I\cup B$ satisfies the body of the corresponding transition feature $r$. 
	
	If the set of next states is empty (i.e., if no transition feature associated with $q$ is satisfied by $I$), the ``else'' branch encodes the different operational semantics that may be defined to govern the automaton's behavior in this case. Typical options for such semantics is either to self-loop on the current state, in which case $\delta(q,I) = \{q\}$, or to reject the input, by moving to an absorbing, ``dead'' state, which we denote by $\bot$. These semantics correspond to two different event consumption policies in CER/F, which are called \emph{skip-till-any-match} and \emph{strict contiguity} respectively \cite{DBLP:journals/vldb/GiatrakosAADG20}.
	
	%Given an MVS $S$ represented by a sequence of interpretations $I_1,\ldots,I_n$ as explained previously, an ASA $\mathcal{A}$ transitions through a state sequence $q_0, \ldots, q_n$, where $q_i = \delta(q_{i-1}, I_i)$. $\mathcal{A}$ accepts $S$ if the last state in the state sequence is an accepting state and it rejects the MVS otherwise.  	
%\end{remark}

\begin{figure*}[t!]\centering
	\includegraphics[width=1\textwidth]{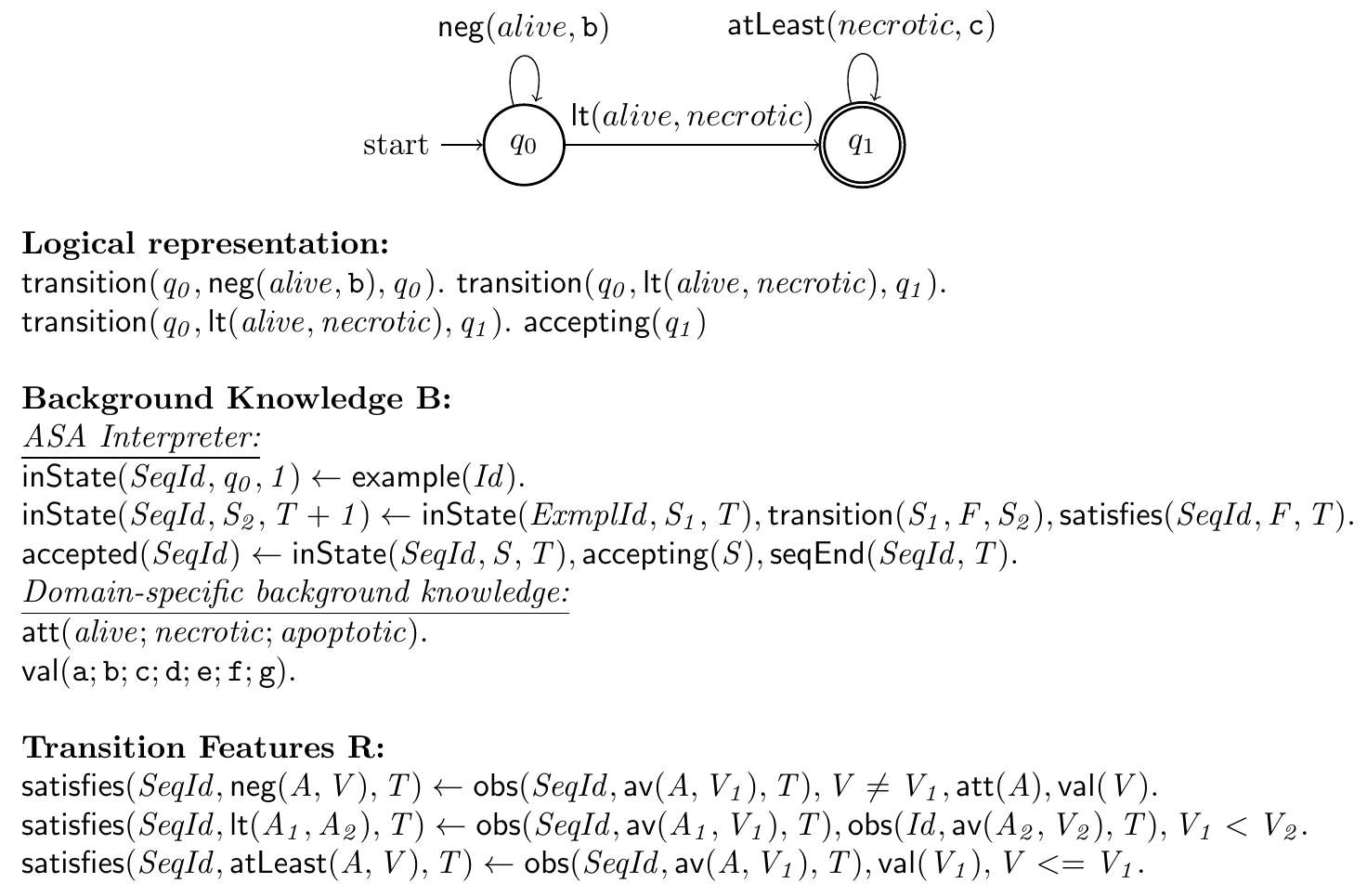}
	\caption{\small An ASA that discriminates between the positive and negative MVSs of Figure \ref{fig:simulation}.}
	\label{fig:asa}
\end{figure*}

\noindent \textbf{The logical representation} of an ASA consists of a set of facts of the form \linebreak $\mathsf{transition}(s_1,f,s_2)$, meaning that the transition from state $s_1$ to $s_2$ is guarded by condition $f$. Figure \ref{fig:asa} presents an example of this representation, which is explained in more detail in Example \ref{ex:asa} below. Additionally, the accepting states in the ASA are specified via atoms of the form $\mathsf{accepting(s)}$.  

\noindent \textbf{Reasoning} with ASA is performed via an interpreter that defines the desired ASA behavior. Such an interpreter is presented in Figure \ref{fig:asa} as part of the background knowledge $B$ (also to be discussed in Example \ref{ex:asa}). The first rule in the interpreter states that for any example (input MVS) $SeqId$ an ASA is initially (at time 1) in state $q_0$, i.e. $q_0$ is the start state. The second rule states that an ASA moves from state $S_1$ to state $S_2$ at time $T$, if there is a transition whose feature is satisfied by the current example ($SeqId$) at time $T$. This is the meaning of the $\mathsf{satisfies}/3$ atom that appears in that rule. Finally, the third rule in the interpreter defines the accepting condition for the MVS $SeqId$, by demanding the ASA to be in an accepting state at the end of the MVS. Rejection of an input MVS is defined implicitly via closed-world assumption. Note that this interpreter implements the strict contiguity operational semantics, i.e. it rejects the input at time $T$ if the set of next states $A_q$ at $T$ is empty. The skip-till-any match semantics may also be supported by adding to the interpreter an extra rule that allows to self-loop when $A_q = \emptyset$, e.g. by using count aggregates in ASP:

$\mathsf{inState}(Id,S,T+1) \leftarrow \mathsf{inState}(Id,S,T), \mathsf{\#count}\{F: \mathsf{satisfies}(Id,F,T)\} = 0.$    

\noindent \textbf{Transition Features.} In this work the transition features are assumed to be defined beforehand (learning them, along with the corresponding ASA is a direction of future work) and are assumed to encode simple relations over attributes and values. Given that the goal is to check if the transition features are satisfied or not by the coordinates of an input MVS, we define them via the predicate $\mathsf{satisfies}/3$ that is used by the interpreter. %, where $F$ is the actual head atom of the transition feature, meaning that the MVS with id $SeqId$ satisfies an instance of $F$ at time $T$. 
The bodies of such $\mathsf{satisfies}/3$ rules consist of the conditions that should be satisfied in order for a transition to be triggered. These conditions are defined by means of attribute-value observation atoms (i.e., $\mathsf{obs}/3$ atoms, as in Table \ref{table:input-data}) and comparison predicates that encode relations between such atoms.  

\begin{example}[\textbf{Answer set automata \& transition features}]
	\label{ex:asa}
	Figure \ref{fig:asa} presents an ASA, its logical representation and its transition features $R$, along with some background knowledge $B$ necessary to reason with the ASA. $B$ consists of the ASA interpreter and the specification of attribute-value domain constants. We first discuss the transition features $R$ before going into the details of the ASA and its functionality. The first rule in $R$ specifies the conditions under which an MVS $SeqId$ satisfies, at time $T$, a predicate with signature $\mathsf{neg}(A,V)$, stating that at time $T$ in $SeqId$ (i.e., in $SeqId$'s $T$-th coordinate), the attribute $A$ does not have the value $V$. The next rule defines a predicate $\mathsf{lt}(A_1,A_2)$ stating that at time $T$ the value of attribute $A_1$ is less than (hence, $\mathsf{lt}$) the value of attribute $A_2$. Finally, the third rule in $R$ defines $\mathsf{atLeast}(A,V)$ stating that at time $T$ the value of attribute $A$ is at least $V$.  
	
	Given these definitions, and based on the instances of the transition features that guard the edges of the ASA in Figure \ref{fig:asa}, the behavior of the ASA is the following: It self-loops on $q_0$ for as long as the size of the alive cells population in the input -- recall our running example -- is not $\mathtt{b}$; it transitions to the accepting state $q_1$ if an observation comes in, where the size of the alive cells population is smaller than that of the necrotic -- an indication that a drug is promising; and it expects the necrotic cells population size to exceed a threshold ($\mathtt{c}$) from that point on, by self-looping on the accepting state.
	
	The ASA in Figure \ref{fig:asa} accepts the positive ($id_1$) and rejects the negative example ($id_2$) in Table \ref{table:input-data}. To see the former note the ASA self-loops on $q_0$ until time $T = 7$, when the value of $\mathit{alive} = \mathtt{b}$, causing $B \cup I_{7}^{id_1}$ to no longer satisfy the self-loop condition on $q_0$. However, at the same time point $\mathit{necrotic} = \mathtt{c}$, which exceeds the value of \emph{alive} (recall that the symbols in the running example represent bins of values and they are ordered by the lexicographic ordering, i.e. letters later in the alphabet represent bins of larger values). Therefore, $B \cup I_{7}^{id_1} \vDash \mathsf{lt}(alive,necrotic)$, causing the ASA to transition to the accepting state $q_1$. The values of \emph{necrotic} in the remaining time steps are at least $\mathtt{c}$, thus the ASA self-loops on the accepting state for the rest of the input.
	
	To see that the ASA rejects $id_2$ note that at time $T=5$ in $id_2$ $\mathit{alive}=\mathtt{b}$, causing the condition that allows to self-loop on $q_0$ up to that point to fail. Moreover, the condition that allows to transition from $q_0$ to $q_1$ is also not satisfied, since $\mathit{necrotic}=\mathtt{b}$. Therefore, since the interpreter implements the strict contiguity semantics, the ASA is in no state at time $T=6$ (i.e. it implicitly moves to the dead state) and the input is rejected. %On the other hand, note that under the skip-till-any match semantics that ASA accepts $id_2$. Indeed, under this semantics the ASA would continue to loop on $q_0$ after $T=5$ until $T=7$, when $\mathit{necrotic}=\mathtt{c}$ and $\mathit{alive}=\mathtt{b}$, yielding $B \cup I_{7}^{id_2} \vDash \mathsf{lt}(alive,necrotic)$ and causing the ASA to move to the accepting state, where it remains until the end of the input.           
\end{example}	

\section{Learning Answer Set Automata}
\label{sec:asal}
We now turn to our approach to Answer Set Automata Learning (\asal) in ASP. Our learning objective may be formulated as follows: Given a training set consisting of positive and negative MVSs $S^+$ and $S^-$, a set of transition features $R$, a ``state budget'' $N$ and potentially, a set of structural and regularization constraints $SC$ and $RC$ respectively, learn an ASA $\mathcal{A}$ that uses the transition features in $R$, respects $SC$, optimizes $RC$, does not exceed $N$ and minimizes the training error, defined as: \linebreak $\sum_{s\in S^{+}} \mathsf{rejects}(\mathcal{A},s) + \sum_{s\in S^{-}} \mathsf{accepts}(\mathcal{A},s)$, i.e. the number of misclassified training examples.

Structural constraints may include e.g. requirements for accepting states being absorbing ones, starting states not being accepting states etc. Regularization constraints are typically related to learning simpler models, where simplicity in our case is measured by the total number of states and transitions in an automaton. Several other simplicity criteria may be considered. For instance, an \emph{earliness bias} would favor automata that accept/reject their input as early as possible, the intuition being that initial segments of the input are often enough to learn a good model, while trying to fit the entire input could yield more complicated automata with inferior generalization capacity. %In contrast, trying to fit the entire input often yields more complicated models with inferior generalization capacity. %As we will see, ASP is an ideal learning framework for defining such types of regularization bias as additional (weak) constraints.

We cast the automata learning problem as an abductive reasoning task in ASP. In its simplest form, such a task may be defined as a tuple $\langle \Pi, IC, G, A \rangle$, where $\Pi$ is a logic program that represents some background knowledge, $IC$ is a set of constraints that must be respected, $G$ is a set of ground constraints, often called ``goals'' and $A$ is a set of predicate signatures called abducibles. A solution to an abductive reasoning task is a set of ground logical facts $\Delta$, such that $B\cup IC \cup \Delta \vDash G$. 

In our case, the background knowledge program $\Pi$ contains the ASA interpreter, the definitions of the transition features $R$ and the training MVSs $S^{+}\cup S^{-}$ in logical form; $IC$ are the structural and regularization constraints, $G$ is a set of ground constraints related to the acceptance/rejection of each MVS in $S^{+}, S^{-}$ respectively and $A = \{\mathsf{transition}/3, \mathsf{accepting}/1\}$. %The latter implies that \asal \ tries to learn a logical specification $\Delta$ of an ASA, as in Figure \ref{fig:asa}, in terms of ground $\mathsf{transition}/3$ and $\mathsf{accepting}/1$ atoms, which accept/reject the training MVSs and satisfy the constraints in $G$ as much as possible, by minimizing the training error.

\begin{table}[t]
	\begin{tabular}{ p{\textwidth} }
		\hline
		%\hline
		\vspace*{-0.8cm}
		\begin{multline*}
		\begin{array}{l}
		\text{\emph{Generate ASA:}} \\
		\{\mathsf{transition}(S_1,F,S_2)\} \leftarrow \mathsf{state}(S_1), \mathsf{state}(S_2), \mathsf{feature}(F). \\
		\mathsf{feature}(\phi) \leftarrow \mathsf{type}_1(var^1_\phi), \ldots, \mathsf{type}_n(var^n_\phi). \\
		\{\mathsf{states}(S)\} \leftarrow \mathsf{maxStates}(S). \\
		\{\mathsf{accepting}(S)\} \leftarrow \mathsf{state}(S). \\
		\mathsf{maxStates}(1..q_{N}). \  \  \ \mathsf{start}(1). \\
		~ \\
		\text{\emph{Minimize the training error:}} \\
		\weak \ \mathtt{accepted}(SeqId),  \ \mathtt{negative}(SeqId). \ [w_{fp}@2,SeqId] \\
		\weak \ not \ \mathtt{accepted}(SeqId),  \ \mathtt{positive}(SeqId). \ [w_{fn}@2,SeqId] \\
		~ \\
		\text{\emph{Example of regularization constraints:}} \\
		\weak \ \mathtt{transition}(S_1,X,S_2). \ [1@1,S_1,S_2,X] \\ 
		\weak \ \mathtt{accepted}(SeqId,T). \ [T@1,SeqId,T] \\
		\mathtt{accepted}(SeqId,T) \leftarrow \mathtt{inState}(SeqId,S,T), \mathtt{accepting}(S).\\
		\mathtt{accepted}(SeqId) \leftarrow \mathtt{accepted}(SeqId,\_).\\
		~ \\
		\text{\emph{Example of structural constraints:}} \\
		\leftarrow \mathtt{transition}(S,_,S2), \mathtt{accepting}, S2 \neq S.
		%\leftarrow \mathtt{start}(X), \mathtt{accepting}(X).
		\vspace*{-0.2cm}
		\end{array}
		\end{multline*}\\
		\hline
	\end{tabular}
	\caption{Abductive ASA learning in ASP.}
	\label{table:gen-test}
\end{table}
\normalsize

The abductive task is straightforward to specify and solve in ASP, via its generate-and-test methodology, according to which we generate automata via choice rules and test their performance via (weak) constraints. Table \ref{table:gen-test} presents an example formulation. The choice rules in the first block of rule generate ASA as collections of $\mathsf{transition}/3$ and $\mathsf{accepting}/3$ facts (as in Figure \ref{fig:asa}). The $\mathsf{feature}/1$ rule is just a ``type rule'' added for each transition feature $r \in R$. For instance, for the transition feature $\mathsf{lt}(A,V)$ from Figure \ref{fig:simulation}, the corresponding type rule is $\mathsf{feature}(\mathsf{lt}(A,V))\leftarrow \mathsf{att}(A),\mathsf{val}(V).$  

The next block of rules is a set of weak constraints that minimize the training error. Note that the  $\mathsf{accepted}/1$ predicate is defined in the ASA interpreter in Figure \ref{fig:asa} and the $\mathsf{positive}/1$ and $\mathsf{negative}/1$ predicates are facts that carry the label of each training MVS. The constraints may be weighted differently, accounting e.g. for imbalances between positive/negative examples in the training set. The weights are $\mathit{w_{fp}}$ for the first rule, which is the price paid for each false positive, and $\mathit{w_{fn}}$ for the second rule, the price paid for each false negative. These constraints have a higher priority, relative to the ones that follow in Table \ref{table:gen-test}, making training error minimization the primary objective.

The next block of rules presents an example of regularization biases in the form of weak constraints. The first constraint attempts to compress the generated automata as much as possible, by penalizing $\mathsf{transition}/3$ facts. The second rule attempts to maximize earliness by minimizing the length of the prefixes that an ASA needs to process before accepting\footnote{Maximizing earliness for input rejection, in addition to acceptance, would also be possible by modifying the ASA interpreter to not handle rejection via the closed-world assumption, but via explicit, absorbing dead states, and adding appropriate regularization constraints.}. The $\mathsf{accepted}/2$ predicate that is used in the earliness constraint simply monitors the number of steps needed to reach an accepting state, while the next rule defines acceptance in terms of such ``partial'' acceptance. To use the earliness bias constraint in \asal \  these two rules should replace the third rule in the ASA interpreter in Figure \ref{fig:asa}, while accepting states should be treated as absorbing. This essentially forces \asal \  to learn an automaton that accepts from prefixes of the input, whose length is to be minimized by the earliness constraint. The last rule in Table \ref{table:gen-test} is a structural constraint ensuring that accepting states are absorbing.

\begin{example}[\textbf{\asal \  in Action}]
	\label{ex:asal-in-action}
	Let $\Pi$ be the program consisting of the rules in Table \ref{table:gen-test}, $B$ and $R$ from Figure \ref{fig:asa} and the logical representation of the two training examples in Table \ref{table:input-data}, as the union of the $I_{t}^{id_1}$'s and the $I_{t}^{id_2}$'s, along with the facts $\mathsf{positive}(id_1)$ and $\mathsf{negative}(id_2)$. Running $\mathsf{Clingo}$ on $\Pi$ and filtering the $\mathsf{transition}/3$ and the $\mathsf{accepting}/1$ facts from the generated solutions yields the learnt ASA. One of these ASA is the one illustrated in Figure \ref{fig:asa}, which is suboptimal, based on the constraints in Table \ref{table:gen-test}. It can be seen that optimal ASA consist of two states and two transitions. e.g. the ASA
	
	$\mathsf{transition}(q_0,\mathsf{at\_least}(alive,\mathtt{e}),q_0). \ \mathsf{transition}(q_0,\mathsf{at\_least}(apoptotic,\mathtt{d}),q_1).$
	
	$\mathsf{accepting}(q_1).$ 
	
	\noindent In addition to being smaller, it can be seen that this ASA accepts the positive example $id_1$ at step $6$, in contrast to the ASA in Figure \ref{fig:asa}, which accepts at step 8. If we drop the earliness constraint in Table \ref{table:gen-test} we may obtain a single-state ASA:
	
	$\mathsf{transition}(q_0,\mathsf{neg}(apoptotic,\mathtt{f}),q_0).$
	
	$\mathsf{accepting}(q_0).$  
	
	\noindent Although this ASA correctly classifies the examples, it is degenerate and has an unconventional behavior, always starting from an accepting state and failing later on to reject the negative examples (e.g. it moves to the implicit dead state at step 5 to reject $id_2$).
\end{example}

\subsection{Incremental Learning \& Automata Revision}

\asal \ is guaranteed to find an optimal, constraint-compliant solution to the abductive ASA learning task, given enough time and memory. The main drawback, however, is that the requirements for such resources grow exponentially with the complexity of the learning task (e.g. alphabet size, number of transition features etc.) and the size of the input (e.g. number and length of the training examples), making the approach infeasible for larger datasets. To alleviate this issue we give-up the requirement for optimal ASA and opt for an incremental learning strategy that simply learns ASA with a good fit in the training data. Additional strategies for improving \asal's scalability, such as incorporating symmetry-breaking constraints, in order to ignore isomorphic ASA, are future work directions.

\asal's incremental learning version operates on mini-batches of the training data, sufficiently small to allow for fast ASA induction from each batch and, ideally, sufficiently large and diverse to allow for learning relatively good ASA from samples of the training set. This is paired with an ASA revision technique, which tries to apply minimal structural modifications on existing automata, to improve their performance on new mini-batches. The algorithm -- we omit the pseudocode due to space limitations -- %, which we denote by $\mathsf{ASAL_{inc}}$ 
is a greedy, iterative hill-climbing search that works as follows: At each point in time, an initially empty, best-so-far ASA $\mathcal{A}$ is maintained. At each mini-batch $D$, if $\mathcal{A}$'s local %$F_1$-score 
classification performance on $D$ is lower than a given error threshold, $\mathcal{A}$ is revised by running $\mathsf{Clingo}$ on a program $\Pi$ similar to the one described in Example \ref{ex:asal-in-action}, augmented as follows: For each $\mathsf{transition}(q_i,\phi_{i},q_j)$ and each $\mathsf{accepting}(q_i)$ fact in the logical specification of $\mathcal{A}$ we add to $\Pi$ the following:

$\mathsf{existing}(\mathsf{transition}(q_i,\phi_{i},q_j)).$

$\mathsf{existing}(\mathsf{accepting}(q_i)).$

$\weak \ \mathsf{not} \ \mathsf{transition}(q_i,\phi_{i},q_j), \mathsf{existing}(\mathsf{transition}(q_i,\phi_{i},q_j)). \ [{-} w_i @ 1]$

$\weak \ \mathsf{not} \ \mathsf{accepting}(q_i), \mathsf{existing}(\mathsf{accepting}(q_i)). \ [1 @ 1]$

\noindent The first two facts simply state which facts already exist in the structural description of $\mathcal{A}$, while the weak constraints penalize their removal, thus fostering minimal revisions. The weight $w_i$ in the first weak constraint is defined as $w_i = n-p$, where $n,p$ are respectively the number of negative/positive examples throughout the entire training set, which are accepted by $\mathcal{A}$ and the acceptance paths use the corresponding transition feature $\phi_i$. Note the ``{-}'' sign in front of $w_i$, which makes $w_i$ positive (i.e., a penalty in the optimization process) if $n > p$ and negative (i.e. a reward) in the opposite case.    

An ASP solver runs on the augmented program $\Pi$ for a given time $t$, and the $k$ best solutions found in that time are preserved, where the time-out $t$ and $k$ are run-time parameters. Subsequently, these $k$ locally best solutions are evaluated on the entire training set (updating the aforementioned $p,n$ counts for each $\mathsf{transition}/3$ fact in these ASA). If an ASA $\mathcal{A}'$ is found after this process has a better global performance than $\mathcal{A}$, it replaces $\mathcal{A}$ as the new best ASA. The process is repeated for a given number of iterations, by shuffling and re-partitioning the training set into new mini-batches at the beginning of each iteration. The current best ASA that results from this process is returned in the output.

%\scriptsize
\begin{table*}[t]
	%\small
	%\footnotesize
	%\scriptsize
	\begin{minipage}{\textwidth}
		\begin{tabular}{>{\centering\arraybackslash}p{1.7cm}>{\centering\arraybackslash}p{1.5cm}>{\centering\arraybackslash}p{2cm}>{\centering\arraybackslash}p{2cm}>{\centering\arraybackslash}p{2cm}>{\centering\arraybackslash}p{2cm}}
			\hline
			\hline
			%\scriptsize
			~ & \textbf{Method} & 
			\textbf{$\mathbf{F_1}$\textbf{-score}} &
			\textbf{\#States} & \textbf{\#Transitions} & \textbf{Time (min)}  \\
			\noalign{\smallskip}
			%\noalign{\smallskip}
			%\hspace*{-2cm}
			
			Bio Small-U                   & \rpni  & 0.702 & 13 & 292 & \textbf{0.05}	  \\
			\noalign{\smallskip}
			~                             & \edsm  & 0.722 & 12 & 278 & \textbf{0.05}  \\
			\noalign{\smallskip}
			~                             & \disc  & 0.833 & 10 & \textbf{13} &  51  \\
			\noalign{\smallskip}
			~                             & $\mathsf{ASAL_{classic}^{batch}}$  & \textbf{0.887} & \textbf{3} & 35 & 1 (time-out)  \\
			\noalign{\smallskip}
			~                             & $\mathsf{ASAL_{classic}^{incr}}$  & 0.882 & \textbf{3} & 41 & 0.566  \\
			
			\noalign{\smallskip}
			\hline
			\noalign{\smallskip}

			Bio Large-U                             & $\mathsf{ASAL_{classic}^{incr}}$  & 0.858 & 3 & 57 & 13  \\
			
			\noalign{\smallskip}
			\hline
			\noalign{\smallskip}
			
			Bio Small                   & $\mathsf{ASAL_{classic}^{batch}}$  & 0.902 & \textbf{3} & 53 & 5 (time-out)  \\
			\noalign{\smallskip}
			~                             & $\mathsf{ASAL_{classic}^{incr}}$  & 0.889 & \textbf{3} & 60 & \textbf{2.7}  \\
			\noalign{\smallskip}
			~                             & $\mathsf{ASAL_{symb}^{batch}}$  & \textbf{0.968} & \textbf{3} & \textbf{5} & 5 (time-out)  \\
			\noalign{\smallskip}
			~                             & $\mathsf{ASAL_{symb}^{incr}}$  & 0.924 & \textbf{3} & 7 & 3.4  \\

			\noalign{\smallskip}
			\hline
			\noalign{\smallskip}
			
			Bio Large & $\mathsf{ASAL_{classic}^{incr}}$  & 0.852 & \textbf{3} & 12 & \textbf{25}  \\
			\noalign{\smallskip}
			~                     & $\mathsf{ASAL_{symb}^{incr}}$  & \textbf{0.942} & \textbf{3} & \textbf{8} & 39  \\
			
			\noalign{\smallskip}
			\hline
			\noalign{\smallskip}
			
			Maritime              & $\mathsf{ASAL_{classic}^{incr}}$  & 0.892 & \textbf{3} & 15 & \textbf{6}  \\
			\noalign{\smallskip}
			~                     & $\mathsf{ASAL_{symb}^{incr}}$  & \textbf{0.952} & \textbf{3} & \textbf{8} & 18  \\
			
			%\noalign{\bigskip}
			\hline
			\hline
		\end{tabular}
		%\vspace{0.5\baselineskip}
	\end{minipage}
	\caption{Experimental results}\label{table:results}
\end{table*}
\normalsize

\section{Experimental evaluation}
We empirically assess \asal \ on two CER datasets from the domains of precision medicine and maritime monitoring. The former has already been outlined in Section \ref{sec:background}. It consists of 644 time series, each containing three signals related to the \emph{alive}, \emph{necrotic} and \emph{apoptotic} attributes, of length 49 each. The positive class (interesting simulations) amounts to the $20\%$ of the data. In addition to this dataset, to which we refer as \emph{Bio-small}, in order to test the scalability of the incremental version of \asal, we also used a significantly larger dataset with the same characteristics, but with 50K training simulations.%, while the remaining $80\%$ corresponds to the negative examples.  

The maritime dataset contains data from nine vessels that cruised around the port of Brest, France. There are five attribute signals for \emph{longitude, latitude, speed, heading}, and \emph{course over ground}. The positive class is related to whether a vessel eventually enters the port of Brest and there are 2,980 negative and 2,269 positive examples, a total of 5,249 multivariate examples, each of length 30. The maritime dataset has been previously used in CER research, we therefore refer to \cite{alevizos2022complex} for more details. Both datasets were discretized using SAX with ten bins.% (i.e. the attributes in the datasets may have up to ten values).  

We compared the following algorithms: (a) $\mathsf{ASAL_{classic}^{batch}}$ and $\mathsf{ASAL_{classic}^{incr}}$, the batch and incremental version of \asal \ that learns ASA that resemble classical automata, by using no transition feature other that equality, i.e. an attribute having a particular value found in the data. This is similar to using a symbol for each attribute-value. This version of \asal \ was evaluated to assess the merits of using relational transition features; (b) $\mathsf{ASAL_{symb}^{batch}}$ and $\mathsf{ASAL_{symb}^{incr}}$, the batch and incremental versions respectively of \asal, as described in the previous sections. These were used with five predefined transition features, similar to those presented in Figure \ref{fig:asa}; (c) \rpni \ \cite{oncina1992identifying} and an improved version thereof, \edsm \ \cite{lang1998results}, two widely used algorithms of the state-merging (SM) family, that compress a PTA (see Section \ref{sec:background}) using greedy heuristics. Note that although these algorithms are quite old, the main ideas behind them are the SoA in SM-style learning and their $\mathsf{LearrnLib}\footnote{\url{https://learnlib.de/}}$ implementation used in the experiments is extremely efficient and frequently used by practitioners; (d) \disc \ \cite{shvo2021interpretable}, a recent algorithm that translates the PTA compression problem into a Mixed Integer Linear Programming problem, which it delegates to the off-the-shelf Gurobi solver. \disc \ is similar to \asal \ in the sense that it is able to learn optimal automata, given enough time and memory.  

The experimental setting was a five-fold cross validation with $80/20$ training/testing set splits. The experiments were carried-out on a 3.6GHz processor (4 cores, 8 threads) and 16GB of RAM. Whenever $\mathsf{ASAL_{classic}^{batch}}$ was used it was given a timeout (maximum allowed time), in order to obtain a solution in a feasible amount of time.

The results are presented in Table \ref{table:results} in the form of average testing set $F_1$-scores, number of states and transitions, as well as training times for the algorithms compared. Note that \rpni, \edsm \ and \disc \ deal with single strings and cannot handle multivariate input. To compare to these algorithms we used a univariate version of the bio dataset, which contains the \emph{alive} attribute only, as it is informative enough to learn a useful model. No such attribute has this property in the maritime dataset, we therefore did not experiment with \rpni, \edsm \ and \disc \ on it. Moreover, only the ``classic'' version of \asal \ was used on the univarate bio dataset, since there are no cross-feature relations that could be captured by transition features in the symbolic version.

The first block in Table \ref{table:results} concerns the small version of the univariate dataset. It can be seen that \rpni \ and \edsm \ are lightning-fast, learning a model in approx. three secs. On the other hand, they have significantly inferior predictive performance as compared to all other algorithms and they are significantly more complicated, as indicated by their size. \disc \ has a better $F_1$-score and learns slightly simpler models, as indicated by its states number (the reduced number of transitions for \disc \ is misleading, since \disc \ omits self loops). % (it operates under the skip-till-any-match semantics). 
On the other hand, it takes a little less than an hour on average to train. $\mathsf{ASAL_{classic}^{batch}}$ has the best predictive performance, achieved within 1 min. Its incremental version closely follows in predictive performance in almost half the time. 

In the large version of the univiariate bio dataset (second block in Table \ref{table:results}) $\mathsf{ASAL_{classic}^{incr}}$ was the only usable algorithm, since \rpni, \edsm \ and \disc \ terminated with memory errors, due to the size of the dataset. In contrast, thanks to its incremental learning strategy that never loads the entirety of the training data into memory, \asal \ was able to learn a good model in approx. 13 minutes.

The results from the small version of the full bio dataset (containing all the features) in the next block in Table \ref{table:results}), highlight the advantages of learning symbolic automata. Indeed, both the batch and the incremental version of \asal \ achieve significantly better results than the classical version of \asal \ and learn automata with much fewer transitions. Note that the $F_1$-scores of the two versions of the batch \asal \ version were achieved with the same time-out value and the incremental versions of \asal \ have comparative training times. Therefore, the symbolic version learns better models without significantly compromising efficiency. The results from the large bio dataset (full, all features used) and the maritime dataset also seem to confirm this claim.     

\section{Conclusions and Future Work}

We presented a methodology for learning symbolic automata where the transition guards are defined via ASP rules, and evaluated our approach on two CER datasets, demonstrating its efficacy. There are several directions for future work, including a more thorough experimental assessement on more datasets and settings, a formal characterization of the expressive power of ASA in relation to common event algebras used in CER, scalability improvements, e.g. via symmetry breaking, and jointly learning 
the transition features definitions, along with the automaton.   

\section*{Acknowledgements}
This work is supported by the project entitled ``ARIADNE - AI Aided D-band Network for 5G Long Term Evolution'', which has received funding from the European Union's Horizon 2020 research \& innovation programme under grant agreement No 871464, and by the project entitled ``INFORE: Interactive Extreme-Scale Analytics and Forecasting'', which has received funding from the European Union's Horizon 2020 research \& innovation  programme under grant agreement No 825070.

%\subsubsection{Acknowledgements} Please place your acknowledgments at
%the end of the paper, preceded by an unnumbered run-in heading (i.e.
%3rd-level heading).

%
% ---- Bibliography ----
%
% BibTeX users should specify bibliography style 'splncs04'.
% References will then be sorted and formatted in the correct style.
%
\bibliographystyle{splncs04}
\bibliography{refs}

\begin{thebibliography}{10}
\providecommand{\url}[1]{\texttt{#1}}
\providecommand{\urlprefix}{URL }
\providecommand{\doi}[1]{https://doi.org/#1}

\bibitem{agrawal2008efficient}
Agrawal, J., Diao, Y., Gyllstrom, D., Immerman, N.: Efficient pattern matching
  over event streams. In: Proceedings of the 2008 ACM SIGMOD international
  conference on Management of data. pp. 147--160 (2008)

\bibitem{alevizos2022complex}
Alevizos, E., Artikis, A., Paliouras, G.: Complex event forecasting with
  prediction suffix trees. The VLDB Journal  \textbf{31}(1),  157--180 (2022)

\bibitem{alevizos2017probabilistic}
Alevizos, E., Skarlatidis, A., Artikis, A., Paliouras, G.: Probabilistic
  complex event recognition: A survey. ACM Computing Surveys (CSUR)
  \textbf{50}(5),  1--31 (2017)

\bibitem{angluin1987learning}
Angluin, D.: Learning regular sets from queries and counterexamples.
  Information and computation  \textbf{75}(2),  87--106 (1987)

\bibitem{angluin2015learning}
Angluin, D., Eisenstat, S., Fisman, D.: Learning regular languages via
  alternating automata. In: Twenty-Fourth International Joint Conference on
  Artificial Intelligence (2015)

\bibitem{argyros2018learnability}
Argyros, G., D'Antoni, L.: The learnability of symbolic automata. In:
  International Conference on Computer Aided Verification. pp. 427--445.
  Springer (2018)

\bibitem{cugola2010tesla}
Cugola, G., Margara, A.: Tesla: a formally defined event specification
  language. In: Proceedings of the Fourth ACM International Conference on
  Distributed Event-Based Systems. pp. 50--61 (2010)

\bibitem{cugola2012processing}
Cugola, G., Margara, A.: Processing flows of information: From data stream to
  complex event processing. ACM Computing Surveys (CSUR)  \textbf{44}(3), ~15
  (2012)

\bibitem{d2017power}
D'Antoni, L., Veanes, M.: The power of symbolic automata and transducers. In:
  International Conference on Computer Aided Verification. pp. 47--67. Springer
  (2017)

\bibitem{demers2006towards}
Demers, A., Gehrke, J., Hong, M., Riedewald, M., White, W.: Towards expressive
  publish/subscribe systems. In: International Conference on Extending Database
  Technology. pp. 627--644. Springer (2006)

\bibitem{demers2007cayuga}
Demers, A.J., Gehrke, J., Panda, B., Riedewald, M., Sharma, V., White, W.M.,
  et~al.: Cayuga: A general purpose event monitoring system. In: Cidr. vol.~7,
  pp. 412--422 (2007)

\bibitem{diao2007sase+}
Diao, Y., Immerman, N., Gyllstrom, D.: Sase+: An agile language for kleene
  closure over event streams. UMass Technical Report  (2007)

\bibitem{drews2017learning}
Drews, S., D'Antoni, L.: Learning symbolic automata. In: International
  Conference on Tools and Algorithms for the Construction and Analysis of
  Systems. pp. 173--189. Springer (2017)

\bibitem{fisman2021inferring}
Fisman, D., Frenkel, H., Zilles, S.: Inferring symbolic automata. arXiv
  preprint arXiv:2112.14252  (2021)

\bibitem{furelos2021induction}
Furelos-Blanco, D., Law, M., Jonsson, A., Broda, K., Russo, A.: Induction and
  exploitation of subgoal automata for reinforcement learning. Journal of
  Artificial Intelligence Research  \textbf{70},  1031--1116 (2021)

\bibitem{giantamidis2021learning}
Giantamidis, G., Tripakis, S., Basagiannis, S.: Learning moore machines from
  input--output traces. International Journal on Software Tools for Technology
  Transfer  \textbf{23}(1),  1--29 (2021)

\bibitem{DBLP:journals/vldb/GiatrakosAADG20}
Giatrakos, N., Alevizos, E., Artikis, A., Deligiannakis, A., Garofalakis, M.N.:
  Complex event recognition in the big data era: a survey. {VLDB} J.
  \textbf{29}(1),  313--352 (2020)

\bibitem{gold1967language}
Gold, E.M.: Language identification in the limit. Information and control
  \textbf{10}(5),  447--474 (1967)

\bibitem{grez2019formal}
Grez, A., Riveros, C., Ugarte, M.: A formal framework for complex event
  processing. In: 22nd International Conference on Database Theory (ICDT 2019).
  Schloss Dagstuhl-Leibniz-Zentrum fuer Informatik (2019)

\bibitem{de2010grammatical}
De~la Higuera, C.: Grammatical inference: learning automata and grammars.
  Cambridge University Press (2010)

\bibitem{lang1998results}
Lang, K.J., Pearlmutter, B.A., Price, R.A.: Results of the abbadingo one dfa
  learning competition and a new evidence-driven state merging algorithm. In:
  International Colloquium on Grammatical Inference. pp. 1--12. Springer (1998)

\bibitem{letort2019physiboss}
Letort, G., Montagud, A., Stoll, G., Heiland, R., Barillot, E., Macklin, P.,
  Zinovyev, A., Calzone, L.: Physiboss: a multi-scale agent-based modelling
  framework integrating physical dimension and cell signalling. Bioinformatics
  \textbf{35}(7),  1188--1196 (2019)

\bibitem{lifschitz2019answer}
Lifschitz, V.: Answer set programming. Springer (2019)

\bibitem{lin2003symbolic}
Lin, J., Keogh, E., Lonardi, S., Chiu, B.: A symbolic representation of time
  series, with implications for streaming algorithms. In: Proceedings of the
  8th ACM SIGMOD workshop on Research issues in data mining and knowledge
  discovery. pp. 2--11 (2003)

\bibitem{lin2007experiencing}
Lin, J., Keogh, E., Wei, L., Lonardi, S.: Experiencing sax: a novel symbolic
  representation of time series. Data Mining and knowledge discovery
  \textbf{15}(2),  107--144 (2007)

\bibitem{maler2017generic}
Maler, O., Mens, I.E.: A generic algorithm for learning symbolic automata from
  membership queries. In: Models, Algorithms, Logics and Tools, pp. 146--169.
  Springer (2017)

\bibitem{muggleton2014meta}
Muggleton, S.H., Lin, D., Pahlavi, N., Tamaddoni-Nezhad, A.: Meta-interpretive
  learning: application to grammatical inference. Machine learning
  \textbf{94}(1),  25--49 (2014)

\bibitem{oncina1992identifying}
Oncina, J., Garcia, P.: Identifying regular languages in polynomial time. In:
  Advances in structural and syntactic pattern recognition, pp. 99--108. World
  Scientific (1992)

\bibitem{pietzuch2003framework}
Pietzuch, P.R., Shand, B., Bacon, J.: A framework for event composition in
  distributed systems. In: ACM/IFIP/USENIX International Conference on
  Distributed Systems Platforms and Open Distributed Processing. pp. 62--82.
  Springer (2003)

\bibitem{schultz2009distributed}
Schultz-M{\o}ller, N.P., Migliavacca, M., Pietzuch, P.: Distributed complex
  event processing with query rewriting. In: Proceedings of the Third ACM
  International Conference on Distributed Event-Based Systems. pp. 1--12 (2009)

\bibitem{shvo2021interpretable}
Shvo, M., Li, A.C., Icarte, R.T., McIlraith, S.A.: Interpretable sequence
  classification via discrete optimization. In: Proceedings of the 35th AAAI
  Conference on Artificial Intelligence (AAAI). pp. 9647--9656 (2021)

\bibitem{smetsers2018model}
Smetsers, R., Fiter{\u{a}}u-Bro{\c{s}}tean, P., Vaandrager, F.: Model learning
  as a satisfiability modulo theories problem. In: International Conference on
  Language and Automata Theory and Applications. pp. 182--194. Springer (2018)

\bibitem{ulyantsev2015bfs}
Ulyantsev, V., Zakirzyanov, I., Shalyto, A.: Bfs-based symmetry breaking
  predicates for dfa identification. In: International Conference on Language
  and Automata Theory and Applications. pp. 611--622. Springer (2015)

\bibitem{wieczorek2017grammatical}
Wieczorek, W.: Grammatical Inference. Springer (2017)

\bibitem{wu2006high}
Wu, E., Diao, Y., Rizvi, S.: High-performance complex event processing over
  streams. In: Proceedings of the 2006 ACM SIGMOD international conference on
  Management of data. pp. 407--418 (2006)

\bibitem{zhang2014complexity}
Zhang, H., Diao, Y., Immerman, N.: On complexity and optimization of expensive
  queries in complex event processing. In: Proceedings of the 2014 ACM SIGMOD
  international conference on Management of data. pp. 217--228 (2014)

\end{thebibliography}
\end{document}